# SFA-UNet: More Attention to Multi-Scale Contrast and Contextual Information in Infrared Small Object Segmentation

Imad Ali Shah *
ᶜDepartment of Electrical Engineering,
College of Electrical and Mechanical
Engineering, National University of
Sciences and Technology, Islamabad,
44000, Pakistan
iali.ee21ceme@student.nust.edu.pk

Fahad Mumtaz Malik
Department of Electrical Engineering,
College of Electrical and Mechanical
Engineering, National University of
Sciences and Technology, Islamabad,
44000, Pakistan
malikfahadmumtaz@ceme.nust.edu.pk

Muhammad Waqas Ashraf
Department of Computer Engineering,
College of Electrical and Mechanical
Engineering, National University of
Sciences and Technology, Islamabad,
44000, Pakistan
washraf.ce22ceme@student.nust.edu.pk

*Abstract—* Computer vision researchers have extensively worked on fundamental infrared visual recognition for the past few decades. Among various approaches, deep learning has emerged as the most promising candidate. However, Infrared Small Object Segmentation (ISOS) remains a major focus due to several challenges including: 1) the lack of effective utilization of local contrast and global contextual information; 2) the potential loss of small objects in deep models; and 3) the struggling to capture fine-grained details and ignore noise. To address these challenges, we propose a modified U-Net architecture, named SFA-UNet, by combining Scharr Convolution (SC) and Fast Fourier Convolution (FFC) in addition to vertical and horizontal Attention gates (AG) into U-Net. SFA-UNet utilizes double convolution layers with the addition of SC and FFC in its encoder and decoder layers. SC helps to learn the foreground-to-background contrast information whereas FFC provide multi-scale contextual information while mitigating the small objects vanishing problem. Additionally, the introduction of vertical AGs in encoder layers enhances the model's focus on the targeted object by ignoring irrelevant regions. We evaluated the proposed approach on publicly available, SIRST and IRSTD datasets, and achieved superior performance by an average 0.75±0.25% of all combined metrics in multiple runs as compared to the existing state-of-the-art methods. The code can be accessed at https://github.com/imadalishah/SFA_UNet

## I. INTRODUCTION

Infrared small object segmentation (ISOS) plays an essential role in a wide range of computer vision applications, that includes early warning systems, night navigation, maritime surveillance, and UAV search and tracking. The importance of ISOS stems from its all-weather working capabilities, long-range detection, and concealment properties. Despite decades of research, ISOS remains a challenging task due to the low contrast, and insufficient information regarding the shape and texture of objects. Additionally, the potential loss of information during high-level semantic feature processing is another issue in ISOS.

Existing approaches to ISOS can be broadly categorized into 1) traditional methods focusing on image processing-based object detection that require prior knowledge about the object, and 2) Deep Learning Architectures (DLA) based methods such as Convolutional Neural Networks and Vision Transformers [1]. Despite the significant success of DLAs in experimental scenarios as compared to traditional methods; they are sensitive to the selection of hyper-parameters, lack generalization, and struggle to perform well in complex real-world scenes [2]. In particular, ISOS DLAs have an inherent issue with regard to the potential loss of small objects and also lack effective utilization of local contrast and global contextual information, due to their black-box nature [3].

A number of empirical and theoretically supported solutions have emerged in the last decade that showed promising results to overcome the challenges of DLAs. Multi-staged and scaled feature learning has been prominently proposed as a potential solution to improve ISOS performance [4]. Attention mechanisms and attention gates (AG) emerged as an effective tool to enhance the representational learning process. They operate by selectively focusing on relevant features and suppressing irrelevant information [5]. Integrating attention mechanisms into DLAs has proven its importance that include object segmentation [6]. Nonetheless, the existing architectures still face loss of small objects as the process goes deeper into DLAs and not considering the contrast information of the object's neighborhood as well as the global contextual information. Resolution of these issues by its provision to the deep models is vital for the robust and accurate ISOS.

In this paper, we proposed a modified U-Net approach, named SFA-UNet (**S**charr convolution (SC) [7] and fast **F**ourier Convolution (FFC) [8] with vertical **A**ttention-gates based **U**-Net architecture). To address the aforementioned challenges to ISOS, SC and FFC are utilized in encoder-decoder of U-Net with the combination of horizontal (encoder-to-decoder) and vertical (encoders-to-encoder) AGs. Such a combination gives the model an enhanced focus to detect target objects while ignoring irrelevant regions. Additionally, SC and FFC enable the model to effectively capture multi-scale contrast and contextual information. The proposed approach resulted in an effective background-to-foreground segmentation and mitigated information loss faced in the deep models.

We evaluated the proposed SFA-UNet on the publicly available SIRST [9] and IRSTD [10] datasets, and demonstrated state-of-the-art results by significantly outperforming existing models. The integration of SC, FFC, and AGs into SFA-UNet consistently improved ISOS performance. This paper not only presents a better but also offer insights into promising approaches for improving existing ISOS tasks. The key contributions are: -

- Utilization of SC and FFC into U-Nets encoder-decoder blocks for effective extraction of multi-scale contrast and context information in terms of ISOS.

- Integration of vertical AGs in encoder layers of U-Net architecture for better information flow and model

---

focus on relevant targeted objects while ignore irrelevant regions such as background noise.

This paper is organized into six sections: Section I mentioned above is the introduction; Section II reviews existing ISOS literature; Section III summarizes the relevant DLA-based techniques; Section IV discusses SFA-UNet components; Section V contains the experimentation; and Section VI ends the paper with a conclusion.

## II. RELATED WORK

### A. Traditional Models

Traditional methods such as Max-mean-Max-medium [11] and morphological operators such as Top-Hat [12] employed filters to extract the target object from the background. Approaches like LCM [13] with variants ILCM [14], TLLCM [15], and MPCM [16] focused on designing salient measures to segment small objects. The IPI model [17] and variants utilized low-rank decomposition to solve the issue by interpreting input as a superposition of low-rank background and sparse target. Even though the proposed models achieved early promising results in the field of ISOS, these methods suffered from low performance in complex scenarios due to dependence upon the prior knowledge of the segmented target object.

### B. Deep Learning Architectures (DLAs)

In contrast to traditional methods, DLAs have become increasingly dominant in object detection, thanks to their experimentally proven robustness and generalization capabilities since the success of CNN by AlexNet [18]. Multi-scale feature learning has been a popular approach to address ISOS challenges. Existing object detection approaches, such as Faster R-CNN [19], SSD [20], RetinaNet [21], and U-Net [22], incorporated multi-scale feature learning to enhance their detection performance. Feature Pyramid Networks (FPN) [23] combined low-and-high resolution with semantically strong and weak features respectively to improve upon ISOS performance. Wang et al [24] balanced the miss detection and false alarm, named MDvsFA, by the effective utilization of conditional GAN with two generators and one discriminator. Dei Y et al introduced Asymmetric Contextual Modulation (ACM) [9] model to enhance the overall network performance and also presented the first public ISOS based dataset, named SIRST, built from real scenes. Liu et al [25] achieved promising results with their introduced multi-head self-attention for ISOS tasks. Recent works have further advanced the ISOS field such as AGPCNet [26] which incorporated attention-guided context blocks and context pyramid modules, and ISNet [10] model with an additional introduction of ISOS-based dataset, named IRSTD, which took object shapes into account. However, the inherent problems of DLAs are still faced due to potential loss of information in deep models, and limitations in capturing both multi-scale contrast and contextual information, which are crucial for robust and accurate detection.

## III. MAKING ISOS SALIENT WITH DLA-BASED CONVENTIONAL TECHNIQUES

### A. Contrast Information – Central Difference Operators

Traditional convolution operators can be sub-optimal for certain tasks, as they treat all input pixels as valid. As a result, researchers have proposed advanced convolution operators. For instance, gated convolution [27] presented a configurable dynamic feature selection process, and partial convolution conditioned the convolution on only valid pixels. Deformable convolution [28] and its subsequent version enhance the modelling of geometric transformations by adding additional offsets and weights learned from the target task. Central difference-based operators such as Prewitt, Sobel, and Scharr are types of convolution operators that captured the details of intrinsic patterns by combination of both intensities and gradients using central difference filters [7].

### B. Global Information Extraction – Expanding Receptive Fields

The ability of a network to capture multi-scale contextual information is crucial for effective ISOS. While local context, in general, is easy to be extracted using convolution; the degree of global information is usually determined by the network's receptive fields. Strategies for expanding receptive fields to extract global information include stacking convolutions, down-sampling layers, and variants of dilated convolutions such as dilation Atrous Pyramid Pooling (ASPP), and hybrid dilated convolution [29][30] and atrous convolutions [31]. FFC performs convolution operations in the frequency domain and can extract image-level information. Approaches like LAMA [32] and FFC-based monocular depth estimation and semantic segmentation [33] have shown success in different applications.

### C. Capture Fine-Grain Details - Focused Object Detection

Attention mechanisms [5] helped in capturing global information by calculating the correlation between individual pixels. Due to their ability to selectively focus on relevant features and suppress irrelevant information, they gained popularity in several computer vision applications including ISOS. Several works incorporated attention mechanisms into DLAs, such as the CBAM [34] and the Squeeze-and-Excitation (SE) block [35]. Attention-Guided Pyramid Feature Fusion [36] integrated attention mechanisms into FPN to adaptively fuse features at different scales. RefineDet [37] incorporated attention mechanisms into the single-shot detection (SSD) framework. MI$^2$T-UNet [38] combined the output of each encoder block to its next encoder block output through AGs. These attention-based methods have shown promising results in enhancing feature representations and improving detection performance.

## IV. METHODOLOGY

### A. Overview of SFA-UNet

SFA-UNet is based on original U-Net with AG architecture with addition of SC and FFC layers utilized in encoder-decoder block, and additional vertical AGs, as shown in Figure 1. Similar to AGs within the encoder-decoder pairs of a U-Net, vertical AGs are incorporated for encoder-encoder layers. SFA-UNet has three encoder-decoder layers to read the input image with dimensions (256, 256, 1) and the depth of different blocks remained; input image (1), encoder (32, 64, 128), middle (256), decoder (128, 64, 32), and output (1) respectively. Each block had double convolution layers (DCL) and employed additional layers of SC-FFC concatenation in between the first and second convolutions layers. Each DCL block comprised of 3 parts; 1) first Convolution (Conv), Batch Normalization (BN) and Rectifier Linear Unit (ReLU) activation, then 2) SC-FFC block, and 3) second Conv-Bn-Relu (or CBR) layers. Encoder DCL ended with max pooling (stride 2) and decoder DCL started with a concatenation layer with no max pooling at the end.

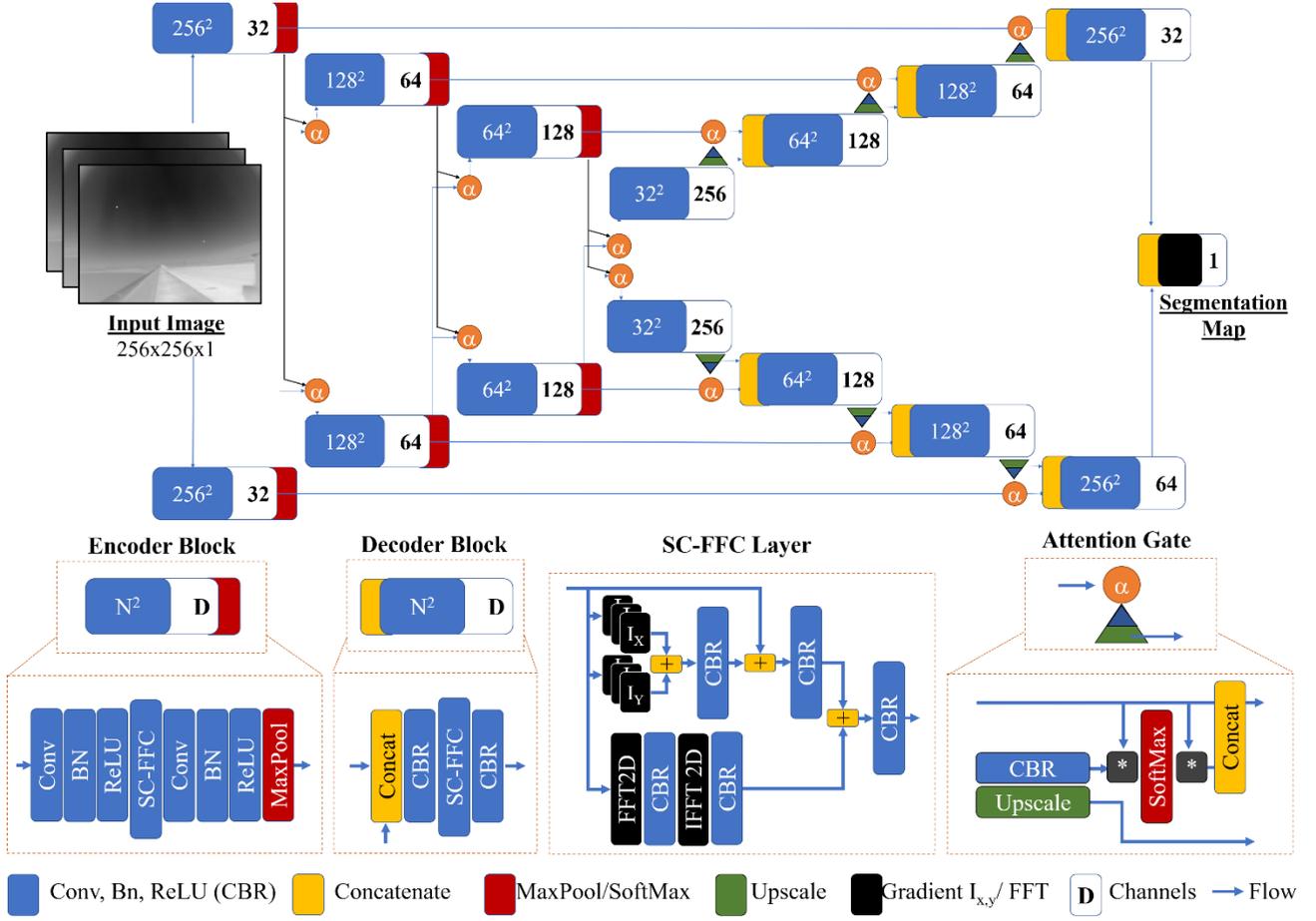

Fig. 1. Proposed SFA-UNet architecture based on U-Net with the integration of SC, FFC and more AGs. Lower half contains building blocks of SFA-UNet with detail of data-path flow and order of different layers, where α, Ix, Iy represent AGs, Schar Operator in x and y direction respectively.

## B. Scharr-fast Fourier Convolutions (SC-FFC) Block

SC is a gradient-based technique for edge detection in images. It employs two distinct kernels, Gx and Gy, to find edges in horizontal and vertical directions. Scharr is more potent than Prewitt and Sobel (modified Prewitt with Gaussian) operators, in terms of detecting edges and enhances image contrast information. Figure 2 shows a comparison results on the SIRST image. FFC transforms images into the frequency domain, enabling faster convolution and efficient for extracting global contextual information, thus remains crucial for tasks like ISOS. FFC uses a larger kernel to gather data from the entire image and does not only rely on local information giving the advantage of in extraction of global contextual information. FFC was utilized as local and global context extraction branch for rich overall information: -

- Acquisition of input (spatial domain), and conversion to the frequency domain by utilizing Real FFT2d and then concatenating the real and imaginary parts: -

$$\text{Real FFT2d}: \mathbb{R}^{H \times W \times C} \to \mathbb{C}^{H \times \frac{W}{2} \times C}$$

$$\text{Complex To Real}: \mathbb{C}^{H \times \frac{W}{2} \times C} \to \mathbb{R}^{H \times \frac{W}{2} \times 2C}$$

- Apply Conv-Bn-ReLU layers in the frequency domain:-

$$Conv \circ Norm \circ Act: \mathbb{R}^{H \times \frac{W}{2} \times 2C} \to \mathbb{R}^{H \times \frac{W}{2} \times 2C}$$

- Convert back to spatial domain from frequency domain by utilizing Real IIFT2d:-

$$\text{Real To Complex}: \mathbb{R}^{H \times \frac{W}{2} \times 2C} \to \mathbb{C}^{H \times \frac{W}{2} \times C}$$

$$\text{Inverse Real FFT2d}: \mathbb{C}^{H \times \frac{W}{2} \times C} \to \mathbb{R}^{H \times W \times C}$$

## C. Vertical and Horizontal Attention Gates

In the modified U-Net with AG, the horizontal AGs retain their original configuration. However, the vertical AGs are set up to operate in dual directions by integration into each encoder blocks. In this arrangement, each block is divided into two parts, and the vertical AGs are positioned in between them. The combination of SC, and FFC into DCLs and in addition to the introduction of cross AGs allows effective focus on the target object and capture multi-scale contrast and contextual information. The modified DCL blocks for encoders and decoders with SC, FFC, and vertical AGs are shown in Figure 1.

TABLE I. PUBLICLY AVAILABLE ISOS-BASED SIRST AND IRSTD DATASETS, AND THEIR USAGE FOR EVALUATION OF SFA-UNET

| Details | SIRST | IRSTD |
|---|---|---|
| Dataset | 427 | 1001 |
| Key Feature | Low Contrast and heavy cluttered objects | More diverse object shapes and complex background |
| Image Size Used | 256 x 256 x 1 (ISOS has single grayscale channel) | |
| Data Split | 8:2 | |

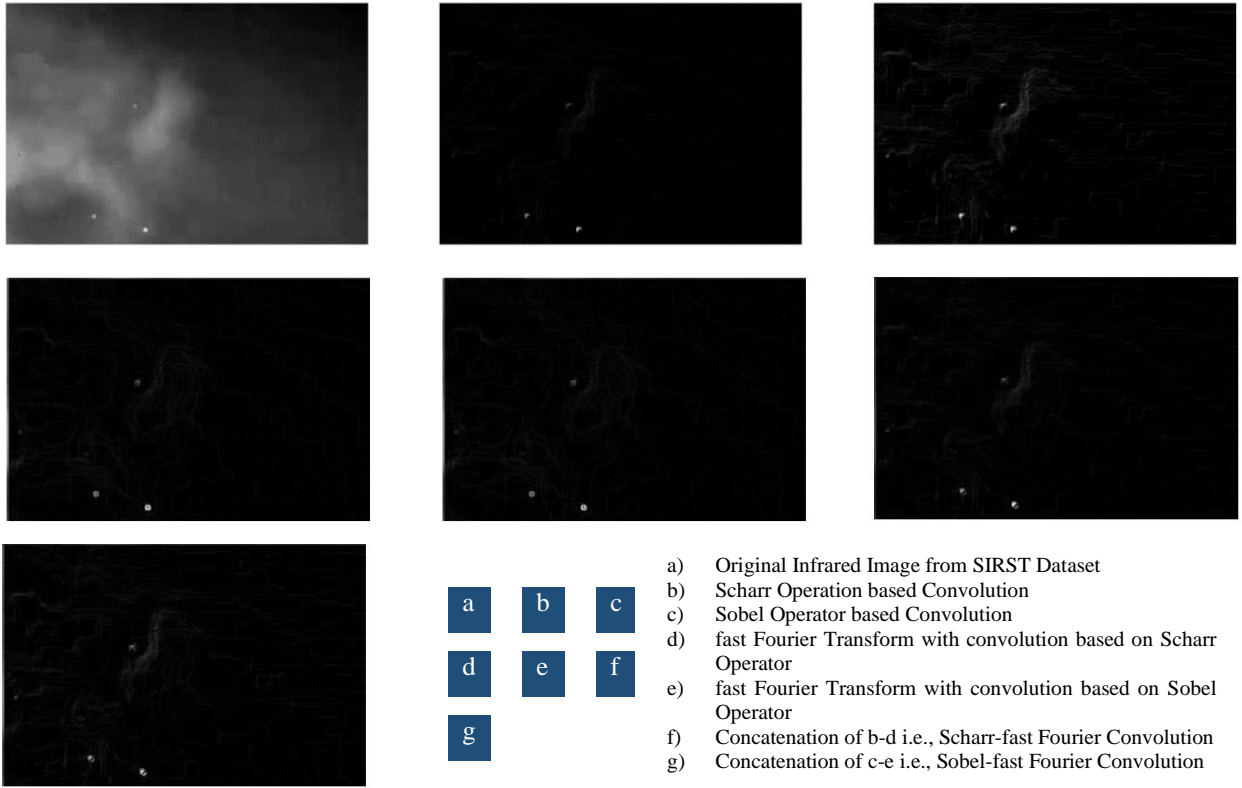

Fig. 2. Qualitative results of the adopted approach being used in SFA-UNet. Results at f-g shows that Scharr-fast Fourier Convolution (SC-FFC) block results with least artifacts i.e., it is better for local and global contrast around the small objects with more clear extraction. (zoom-in for better visualization)

a) Original Infrared Image from SIRST Dataset
b) Scharr Operation based Convolution
c) Sobel Operator based Convolution
d) fast Fourier Transform with convolution based on Scharr Operator
e) fast Fourier Transform with convolution based on Sobel Operator
f) Concatenation of b-d i.e., Scharr-fast Fourier Convolution
g) Concatenation of c-e i.e., Sobel-fast Fourier Convolution

## V. EXPERIMENTATION AND FINDINGS

### A. Datasets

We used publicly available SIRST [9] and IRSTD [10] datasets that are valuable resources for ISOS research, details are shown in Table 1. SIRST is a single-frame dataset of infrared images with low contrast between foreground and background. Whereas, IRSTD provides a wider range of target shapes and detailed annotations. Both datasets serve as valuable benchmarks for advancing ISOS development and are developed for ISOS tasks.

### B. Metrics

In line with other literature on ISOS, we evaluated our proposed approach using pixel-level metrics (IoU and nIoU), object-level (Pd and Fa), and model-level (F-Score and AUC). Formulas for Intersection over Union (IoU), normalized IoU (nIoU), probability of detection (Pd), and false alarm rate (Fa) are: -

$$IoU = \frac{1}{n} \cdot \frac{\sum_{i=0}^{n} tp_i}{\sum_{i=0}^{n}(fp_i + fn_i - tp_i)} \quad Pd = \frac{1}{n} \cdot \sum_{i=0}^{n} \frac{N_{pred}^i}{N_{all}^i}$$

$$nIoU = \frac{1}{n} \cdot \sum_{i=0}^{n} \frac{tp_i}{fp_i + fn_i - tp_i} \quad Fa = \frac{1}{n} \cdot \sum_{i=0}^{n} \frac{P_{false}^i}{P_{all}^i}$$

Where n, tp, fp, and fn stands for total number of samples, true positive, false positive, and false negative, respectively. $P_{false}$, $P_{all}$ stands for the pixels of falsely identified objects and the pixels of all objects, and $N_{pred}$, $N_{all}$ stands for the number of correctly detected objects and the total number of objects.

### C. Implementation Details

We carried out our studies utilizing freely accessible resources of Google Colaboratory [39] and the model was implemented in TensorFlow. We utilized binary cross-entropy loss [40] as our criterion and AdamW optimizer with combination of 0.001 and 0.004 as it's initial learning and weighted decay rates [41] respectively. Each experiment had a batch size of 8, and a maximum epoch of 150 was used to train it.

### D. Quantitative and Qualitative Comparisons

The qualitative results achieved by the components used in SFA-UNet are shown in Figure 2, where its promising effects can be validated. Furthermore, the quantitative performance of SFA-UNet as compared to other selective top performing conventional and DLAs based methods are given in Tables 2-3, respectively. It is clear from these results that SFA-UNet has better demonstratable performance on both ISOS datasets. The improvement of SFA-UNet's performance can be attributed to its modifications discussed in Section IV, allowing a better fusion of feature extraction and information-sharing across multi-scales.

TABLE II. COMPARISON OF MODEL LEVEL PERFORMANCE OF SFA-UNET WITH OTHER STATE-OF-THE-ART MODEL FOR ISOS

| Approach | SIRST | | IRSTD | |
|---|---|---|---|---|
| | F-Score | AUC | F-Score | AUC |
| LCM | 12.80 | 0.058 | 8.52 | 0.099 |
| IPI | 57.63 | 0.448 | 25.17 | 0.248 |
| ACM | 84.02 | 0.684 | 77.59 | 0.719 |
| AGPC | 84.85 | 0.765 | 79.73 | 0.734 |
| UCFNet | 89.43 | 0.843 | 81.60 | 0.745 |
| MI²T-UNet | 88.43 | 0.948 | 84.66 | 0.862 |
| **SFA-UNet** | **89.98** | **0.956** | **86.46** | **0.883** |

TABLE III.  COMPARISON OF PIXEL AND OBJECT LEVEL PERFORMANCES (ROUNDED TO 1-DECIMAL POINT FOR VISUALIZATION)

| Approach | SIRST | | | | IRSTD | | | |
|---|---|---|---|---|---|---|---|---|
| | IoU | nIoU | Pd | Fa | IoU | nIoU | Pd | Fa |
| LCM | 6.8 | 9 | 77.1 | 183.2 | 4.5 | 4.7 | 57.6 | 66.6 |
| IPI | 40.5 | 51 | 91.8 | 148.4 | 14.4 | 31.3 | 86.4 | 450.4 |
| ACM | 72.5 | 72.2 | 93.5 | 12.4 | 63.4 | 60.8 | 91.6 | 15.3 |
| AGPC | 73.7 | 72.6 | 98.2 | 17 | 66.3 | 65.2 | 92.8 | 13.1 |
| UCFNet | 80.9 | 78.9 | **100** | **2.3** | 68.9 | 69.3 | 93.6 | **11** |
| MI$^2$T-UNet | 82.1 | 78.7 | 98.6 | 4.2 | 69.5 | 67.7 | 94.3 | 13 |
| **SFA-UNet** | **83.9** | **79.5** | 99.1 | 2.5 | **69.8** | **69.4** | **94.5** | 11.2 |

## VI. CONCLUSION

In this paper, we identified the prevailing issues in ISOS that affect model accuracy and performance. Inspired by these challenges, we proposed SFA-UNet, a modified U-Net with attention gate (AG) based architecture. SFA-UNet contains Scharr and fast Fourier based convolutions (SC-FFC) in encoder-decoder blocks and vertical AGs within-encoder block. The proposed model effectively addressed the inherited issues in ISOS-based DLA approaches. Specifically, to address the issues of insufficient information in small objects and potential loss in deep models, SC-FFC layers helped to capture multi-scale contrast and contextual information. To deal with the issue of effective extraction of fine-grained details while ignoring potential background noise, addition of vertical AGs within the respective encoder layers of U-Nets, enhanced the model's focus on the targeted object and ignored irrelevant regions. Experimentations validated that the proposed SFA-UNet model demonstrate significant performance and produced an average 0.75±0.25% of all combined metrics in multiple runs as compared to the existing state-of-the-art approaches on both datasets. Moreover, insights from SFA-UNet can be utilized for ongoing ISOS research in infrared based applications, such as remote sensing and surveillance etc.